\documentclass[runningheads]{llncs}
\usepackage{graphicx}
\usepackage{amssymb}
\usepackage{extarrows}
\usepackage{dsfont}

\usepackage[table]{xcolor}
\usepackage{multirow}

\usepackage{amsmath}
\DeclareMathOperator*{\argmin}{argmin}

\def\otmap{\emph{OTMapOnto }}

\newcommand{\R}{\mathbb{R}}
\newcommand{\emb}{\mathbf}

\newcommand{\comment}[1]{}

\def\corresp{\mbox{$\rightsquigarrow$}}

\newcommand{\entity}[1]{\mbox{\textsf{#1}}}

\begin{document}
\title{Exploring Wasserstein Distance across Concept Embeddings for Ontology Matching}
\titlerunning{Wasserstain Distance for Ontology Matching}
%
\author{Yuan An
	\and
Alex Kalinowski
	\and
Jane Greenberg
}
\authorrunning{Yuan An, Alex Kalinowski, and Jane Greenberg}
%
\institute{College of Computing and Informatics, Drexel University, Philadelphia, PA, USA \\
	\email{\{ya45,ajk437,jg3243\}@drexel.edu} 
}

\maketitle              

\begin{abstract}
Measuring the distance between ontological elements is fundamental for ontology matching. 
String-based distance metrics are notorious 
for shallow syntactic matching.
In this exploratory study, we investigate Wasserstein distance targeting continuous
space that can incorporate various types of information. 
We use a pre-trained word embeddings system to
embed ontology element labels. We examine the effectiveness of Wasserstein 
distance for measuring similarity between 
ontologies, and discovering and refining matchings between individual elements. 
Our experiments with
the OAEI conference track and MSE benchmarks achieved competitive results compared to
the leading systems.

\keywords{ontology matching  \and optimal transport \and Wasserstein Distance 
	\and ontology embedding.}
\end{abstract}
%
%
%
%
%
\section{Introduction}
\label{sec:introduction}

Semantic and structural heterogeneity is widespread among ontologies. To bridge the heterogeneity, 
almost all ontology matching systems 
\cite{ontology-matching-literature-review,AML-ontology-matching,LogMap}
need to evaluate the distances between ontological elements.
String-based distance metrics have dominated the field \cite{string-similarity-ontology-matching}. 
However, string-based distance metrics are notorious for shallow syntactic matching.
It is also challenging to determine which string similarity measures to use and how
to effectively combine them in matching systems \cite{string-similarity-ontology-matching}. 
Aligning elements across different ontologies essentially involves measuring the semantic similarity/distance
of the ontological elements representing items in the underlying domains. 
Recently, word embeddings \cite{word2vec} have been
used to successfully encode syntactic and semantic word relationships.
As a result, word embeddings have displayed excellent performance in 
applications for cross-lingual word alignment
\cite{unsupervised-alignment-wasserstein}.
Ontology matching techniques have also been created using embeddings, 
with embedding vectors predominantly utilized as inputs in supervised 
\cite{ontology-alignment-embedding-random-forest,BERTMap}
or distantly supervised \cite{augmenting-ontology-alignment} machine learning models.

While machine learning is effective in making use of ontology embeddings, 
a significant effort is required to gather training instances. 
Although some systems directly use
the cosine similarity between embedding vectors  
\cite{biomedical-ontology-alignment} for deriving candidate matchings, 
the embeddings are first retrofitted through tailored training instances. 
In practice, the present top matching algorithms are largely unsupervised 
and rely only upon existing ontology and external sources.
Motivated by the desired property of unsupervised learning, we posit the following question:   
\emph{If ontology elements can be readily encoded as embedding vectors,   
can we use them in an unsupervised fashion for ontology matching?}

To research the question, we formulate ontology matching
as an optimal transport problem from a source ontology embedding space to 
a target ontology embedding space. 
Optimal Transport (OT) \cite{ot-vallani} 
has been applied to various alignment applications 
including word embeddings alignment
\cite{Gromov-W}, sequence-to-sequence learning \cite{chen2019improving},
heterogeneous domain alignment \cite{got}, and graph comparison and matching 
\cite{an-optimal-graph-comparison}.
A desired advantage of OT-based approach is \emph{unsupervised learning}.
The OT solution establishes
optimal mappings and
a shape-based distance called \emph{Wasserstein distance} between distributions. 
In this study, we explore the effectiveness of 
Wasserstein distance for ontology matching. 
Our inquiry focuses on answering 
the following 3 questions:
\begin{enumerate}
	\item \emph{How effective is Wasserstein distance for measuring similarity
	between (blocks of) ontologies?} 
	\item \emph{How effetive is the coupling matrix accompanying a Wasserstain distance for 
	deriving alignments between individual ontological elements?} 
	\item \emph{How effective is Wasserstain distance for refining matching candidates?} 
\end{enumerate}  
  
The rest of the paper presents our research and is organized as follows.
Section \ref{sec:optimal-transport} introduces optimal transport and 
Wasswerstein distance. 
Section \ref{sec:ontology-embeddings-matching} describes the ontology embeddings used in this study. 
Section \ref{sec:ontology-wasserstein-distance} measures ontology similarity 
in Wasswerstein distance.
Section \ref{sec:coupling-matrix-matching} derives matching candidates. 
Section \ref{sec:refining-matching} refines matching candidates. 
Section \ref{sec:experimental-results} presents our experiment and results. 
Section \ref{sec:discussion} comments on the results.
Section \ref{sec:related-work} discusses related work, and  
finally, Section \ref{sec:conclusion} concludes the paper with future directions.

%
%
%

\section{Optimal Transport and Wasserstein Distance} 
\label{sec:optimal-transport}

Optimal transport (OT) \cite{ot-vallani}
originated as a solution to  
the problem of transporting masses from one configuration onto 
another with \emph{the least effort}. OT had deep connections with 
major economic problems, for example, in logistics, production planning, and network routing, etc. Since then,  
optimal transport has been generalized from practical concerns to powerful mathematical tools
for comparing distributions. In particular, given 
two point sets modeled as two discrete distributions, 
optimal transport is an effective approach
for discovering a minimum cost mapping between the two sets. 
Considering the set of embedding vectors of an ontology as a set of points, 
we can apply optimal transport to discover a minimum cost mapping between two sets of ontology embeddings.

Formally, given two sets of ontology embeddings $\mathbf{X}=\{\emb{x}_{i}\in \R^{d}, i=1..n\}$ and 
$\mathbf{Y}=\{\emb{y}_{j}\in \R^{d}, j=1..m\}$,
where each embedding is represented as a vector  $\emb{x}_{i}$ or $\emb{y}_{j}\in \R^{d}$. Let
$\mu = \sum_{i=1}^{n}{p}(\emb{x}_{i}) \delta_{\emb{x}_{i}}$
be the probability distribution defined on the set $\mathbf{X}$, where
$\delta_{\emb{x}_{i}}$ is the Dirac at the point $\emb{x}_{i}$ and 
$p(\emb{x}_{i})$ is a probability weight
associated with the point $\emb{x}_{i}$.
Similarly, let $\nu = \sum_{j=1}^{n}{q}(\emb{y}_{j}) \delta_{\emb{y}_{j}}$
be the probability distribution defined on the set $\mathbf{Y}$, where
$\delta_{\emb{y}_{j}}$ is the Dirac at the point $\emb{y}_{j}$ and 
$q(\emb{y}_{j})$ is a probability weight
associated with the point $\emb{y}_{j}$.
Usually, we consider uniform weights, e.g., $p(\emb{x}_{i}) = \frac{1}{n}$, for $i=1..n$, 
and $q(\emb{y}_{j}) = \frac{1}{m}$, for $j=1..m$. However, if additional information
is provided, $p(\emb{x}_{i})$ and $q(\emb{y}_{j})$ can incorporate the information as non-uniform distributions. 
Optimal transport (OT)
defines an optimal plan for mass transportation and 
a distance between the two distributions. 

Specifically, let $\mathbf{C}=[c(\emb{x}_{i}, \emb{y}_{j})]_{i, j}$ be 
a ground cost matrix with $c(\emb{x}_{i}, \emb{y}_{j})$ measuring a ground distance between the individual 
embeddings $\emb{x}_{i}$ and $\emb{y}_{j}$. 
Let $\mathbf{T}=[T(\emb{x}_{i}, \emb{y}_{j})]_{i, j}$ be 
a matrix of a transport plan (or \emph{couplings}) 
with $T(\emb{x}_{i}, \emb{y}_{j})$ specifying how much mass will be transported from point 
$\emb{x}_{i}$ to point $\emb{y}_{j}$. Let $\Pi(\mu, \nu)$ be the set of all feasible transport plans defined as:
\begin{equation}
    \label{eq:transport-plan}
    \Pi(\mu, \nu)\stackrel{\text{def}}{=}\{\mathbf{T}\in \R_{+}^{n\times m}| \mathbf{T} \mathds{1}_{n}=\mu, 
    \mathbf{T}^{\top} \mathds{1}_{m}=\nu\} 
\end{equation}
where $\mathds{1}_{n}$ and $\mathds{1}_{m}$ are all one vectors, 
$\mathbf{T} \mathds{1}_{n}=\mu$ and 
$\mathbf{T}^{\top} \mathds{1}_{m}=\nu$ are marginal constraints on feasible plans. 
The \emph{Optimal Transport} problem is to find the map $\mathbf{T}: \mathbf{X}\rightarrow \mathbf{Y}$, 
where  
\begin{equation}
	\label{eq:optimaltransport}
	\mathbf{T} = \argmin_{\mathbf{T}\in \Pi(\mu, \nu)} \sum_{i=1}^{n}\sum_{j=1}^{m} c(\emb{x}_{i}, \emb{y}_{j})\cdot T(\emb{x}_{i}, \emb{y}_{j}), 
	\text{s.t.}, \mathbf{T} \mathds{1}_{n}=\mu, \mathbf{T}^{\top} \mathds{1}_{m}=\nu
\end{equation}
The map $\mathbf{T}$ is also called \emph{coupling matrix}. It gives rise to a distance measure between the two 
distributions called \emph{Wasserstein distance} defined as:
\begin{equation}
	\label{eq:wasserstein}
	W(\mu, \nu)\stackrel{\text{def}}{=}\min_{\mathbf{T}\in \Pi(\mu, \nu)} \langle \mathbf{C}, \mathbf{T} \rangle
	{=}\min_{\mathbf{T}\in \Pi(\mu, \nu)} \sum_{i=1}^{n}\sum_{j=1}^{m} c(\emb{x}_{i}, \emb{y}_{j})\cdot T(\emb{x}_{i}, \emb{y}_{j})
\end{equation}
The optimization problem can be efficiently solved by 
replacing the objective with an entropy regularized objective 
such as in the sinkhorn algorithm \cite{ot-vallani}.

\begin{figure}[!ht]
	\centering
	\includegraphics[width=1\textwidth]{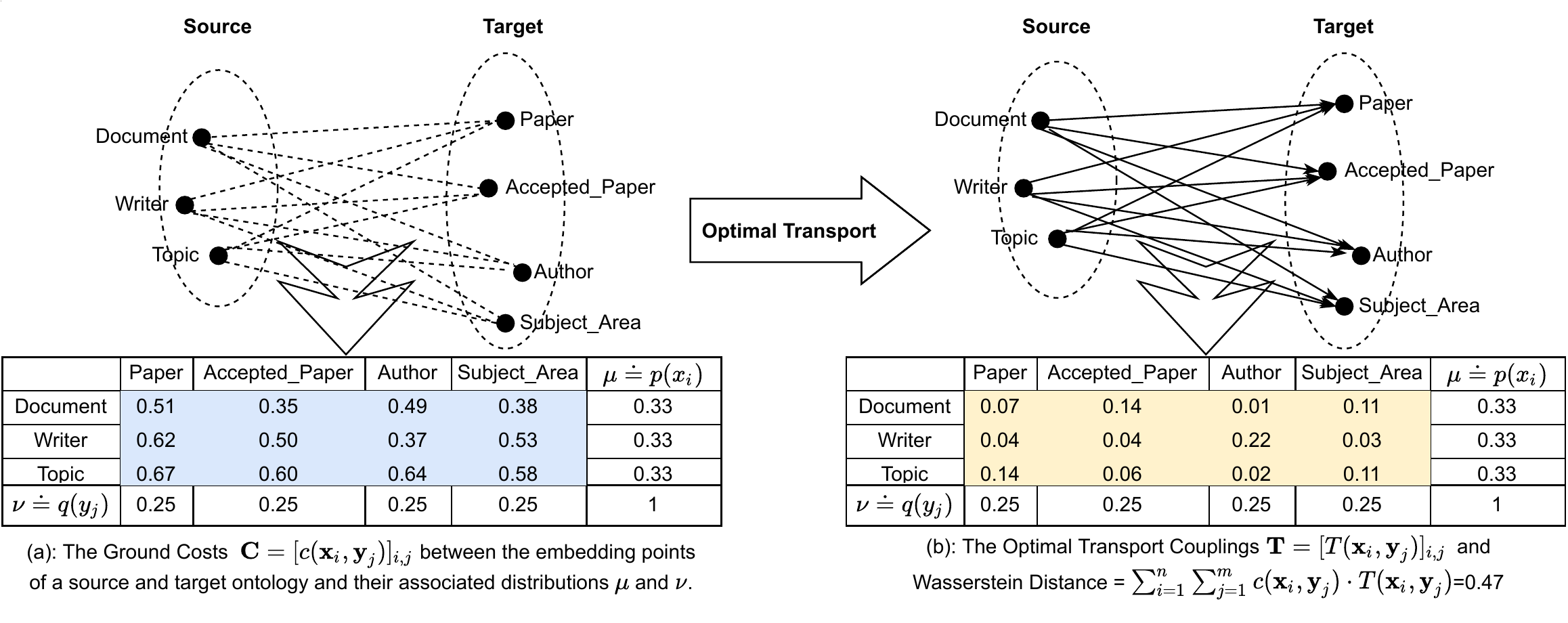}
	\caption{Optimal Transport Problem between a Source and Target Ontology}
	\label{fig:OT-example}
\end{figure}

\example{\emph{
Figure \ref{fig:OT-example} illustrates the configuration of the optimal transport problem 
between a source and target ontology. The source ontology has 3 concepts: \entity{Document}, 
\entity{Writer} and \entity{Topic}. The target ontology has 4 concepts which are: \entity{Paper}, 
\entity{Accepted\_Paper}, 
\entity{Author} and \entity{Subject\_Area}. The table in Figure \ref{fig:OT-example} (a) shows
the Euclidean distances as the ground costs 
between the embeddings of the source and target ontology concepts. The points in both the source and target 
embedding sets are uniformly distributed, as reported in the last column $\mu$ and the last row 
$\nu$. By solving Equation (\ref{eq:optimaltransport}), the table in Figure \ref{fig:OT-example} (b)
displays the optimal transport couplings between the points in the source and target embedding spaces.
Notice the couplings are feasible because they satisfy the following marginal constraints: $\mathbf{T} \mathds{1}_{n}=\mu$ and 
$\mathbf{T}^{\top} \mathds{1}_{m}=\nu$. 
By solving the Equation (\ref{eq:wasserstein}), 
the Wasserstein distance between the two embedding sets is 0.47
given the ground costs and the couplings.   
}}
\label{ex:OT-example}

%
%
%

\section{Ontology Embeddings and Matching}
\label{sec:ontology-embeddings-matching}

Ontology embedding is the problem of encoding 
ontological elements as numerical vectors.
Various methods have been proposed for representing individual components in an ontology as embeddings. 
For example, translational-based methods \cite{transe,transr} and 
graph neural networks (GNN) \cite{graph-representation-learning} encode an ontology based on its
graph structure. Text-enhanced methods for ontology embeddings 
\cite{utilizing-textual,unsupervised-embedding-enhancements,an-etal-2018-accurate,model-text-enhanced,Wang2016TextEnhancedRL} 
encode lexical words of ontology elements. Logic-aware methods \cite{rocktaschel-etal-2015-injecting,demeester-etal-2016-lifted}
incorporate logical constraints into ontology embeddings. 
It is attempting to encode ontologies using the above methods and 
directly apply optimal transport for discovering mapping. 
However, 
simply applying these embeddings for ontology matching is negatively impacted by   
the \emph{lack of registration} problem \cite{pmlr-v108-alvarez-melis20a}. Specifically, 
these embeddings were mainly developed for applications concerned with a single ontology,
for example, link prediction \cite{open-challenge-inductive-link-prediction}. 
The embedding spaces of two independent ontologies may mismatch due to different dimensions or 
various rotations and translations. As a result, 
there may not exist a distance or 
the direct geometric locations between the points in the embedding spaces may not reflect their
underlying genuine relationships. This is a significant issue for optimal transport-based
approach which needs a meaningful ground cost. 

We will study the problem of matching the embeddings incorporating structural and logical
information in future work.
In this exploratory study, we focus on 
the ontology embeddings corresponding to 
the labels of the ontology elements, for example, the \emph{`rdfs:label'} of 
a concept.  
we apply the pre-trained language model, \emph{fasttext} \cite{fasttext2016enriching}, 
to encode the labels of the set of ontology concepts, object and 
datatype properties.
Using the same pre-trained language model to encode the labels of different ontologies will alleviate
the lack of registration problem, because the resultant embeddings are 
in the same embedding space. We will show that
OT on label embeddings already produce promising results.
For each element, we first normalizes the element's 
label via a sequence of standard text processing steps. If necessary, 
the labels are augmented with synonyms. 
We then split the normalized label
into individual words which in turn are fed into the pre-trained language model to
obtain their corresponding word embeddings. 
For the element, we obtain its embedding by computing the average of the set of embeddings or
use the entire set of the embeddings of the individual words for next-step processing.

%
%
%
\section{Wasserstain Distance for Measuring Ontology Similarity}
\label{sec:ontology-wasserstein-distance}

Prior to applying optimal transport and Wasserstein distance for ontology matching, 
we first analyze some properties of Wasserstein distance for capturing ontology similarity.
A desirable property is that Wasserstein distance should closely correlate with ontology similarity. 
That is, the more similar two ontologies are, the shorter the Wasserstein distance 
between them is. Quantitatively measuring the similarity between two ontologies is a very challenging problem.
One option is to count 
the minimum number of graph edit operations to transform one graph to another.
It is known that graph edit distance is NP-hard and it depends on a set of graph edit operations.
Another option is the Jaccard Index on sets.  
If matchings between ontologies are available,
we can define the Jaccard similarity between ontologies as follows:
\[
Jac(\mathcal{O}_{S}, \mathcal{O}_{T}) = \frac{|M|}{|\mathcal{O}_{S}|+|\mathcal{O}_{T}|-|M|},
\] 
where $\mathcal{O}_{S}$ and $\mathcal{O}_{T}$ are source and target ontologies, 
$|\mathcal{O}_{S}|$ and $|\mathcal{O}_{T}|$ return the number of concepts in each ontology,
$M$ is the set of matchings between $\mathcal{O}_{S}$ and $\mathcal{O}_{T}$, and 
$|M|$ gives the number of matchings.  

\begin{figure}[t]
	\centering
	\includegraphics[width=.7\textwidth]{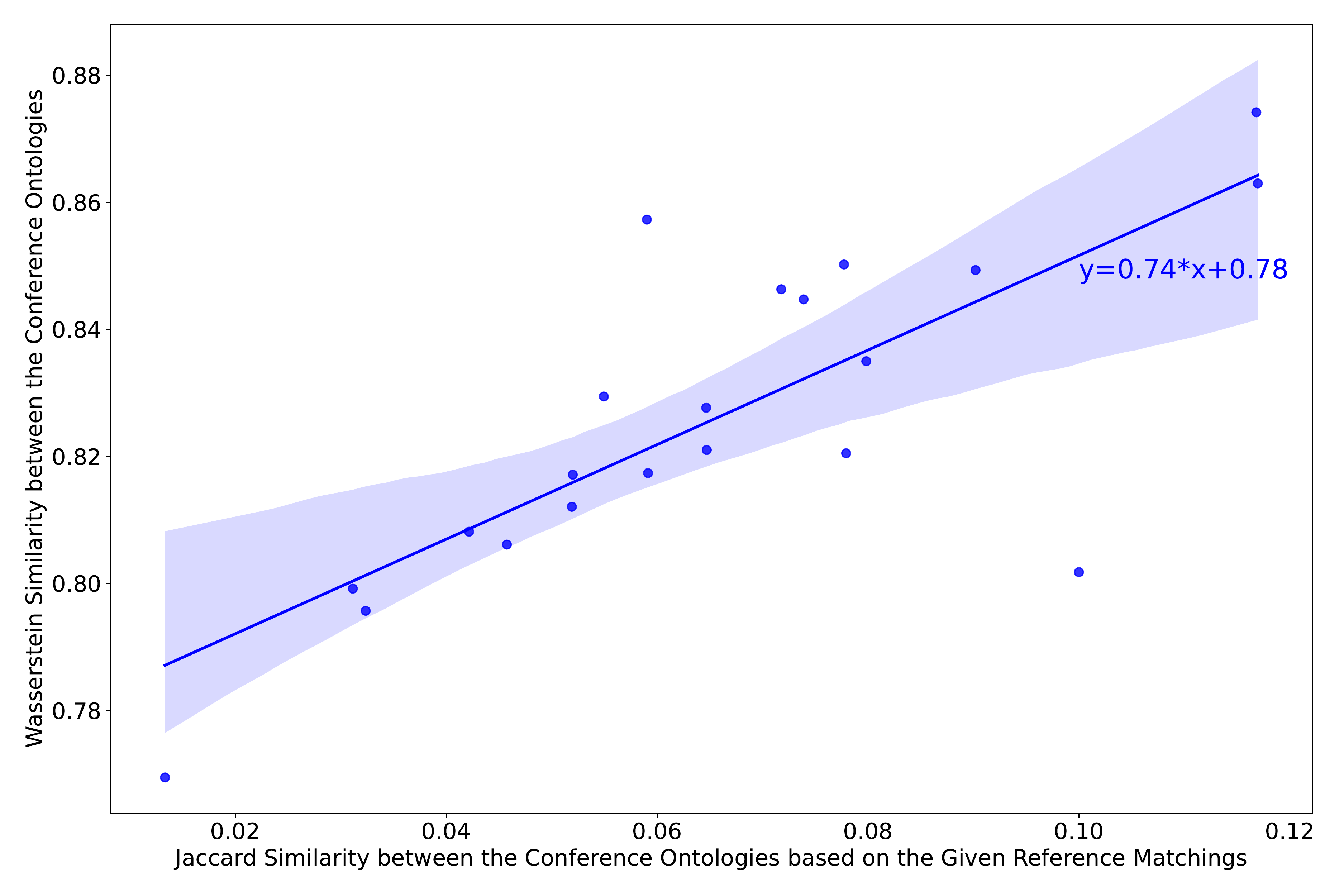}
	\caption{Regression plot between Wasserstein similarity ($e^{-wd}$) and Jaccard similarity based on the given
		matching references for the conference ontologies}
	\label{fig:wd_jac_regplot}
\end{figure}

The OAEI Campaign provides a list of ontology matching tasks each with a set of curated matching references. 
In particular, the Conference track contains 21 matching cases among 7 conference ontologies.
For each case, we compute: (1) its Jaccard similarity based on the given matching references, and
(2) the Wasserstein distance between the ontology embeddings. We convert a Wasserstein distance, $wd$, to
a Wasserstein similarity, $ws$ as $ws = e^{-wd}$.  
Figure \ref{fig:wd_jac_regplot} shows the
regression plot between the Wasserstein similarities and Jaccard similarities. 
Furthermore, we calculated the Pearson correlation coefficient (PCC)
between the two sets of similarities defined as 
$\rho = \frac{cov(X,Y)}{\sigma_X \sigma_Y}$. 
The PCC is 0.77, indicating 
Wasserstein distance is highly correlated with the Jaccard coefficients when matchings between 
two ontologies are known. As a result, Wasserstein distance exhibits the desirable property for capturing ontology similarity.  
We can then leverage this good property for developing 
algorithms that derive and refine matchings using optimal transport as 
presented below.

%
%
\section{Deriving Matching Candidates from Global Coupling Matrix}
\label{sec:coupling-matrix-matching}

Given a source $\mathcal{O}_{S}$ and a target $\mathcal{O}_{T}$ ontology, 
we first apply optimal transport \emph{globally} to the ontologies for 
generating a set of candidate matchings. We then compute contextual  
\emph{local} Wasserstein distances to  
refine the matchings by filtering out false positives. 

In this section, we describe the building blocks for deriving matching candidates globally.  
Let $\mathbf{X}=\{\emb{x}_{i}\in \R^{d}, i=1..n\}$ be the source ontology concept embeddings and 
$\mathbf{Y}=\{\emb{y}_{j}\in \R^{d}, j=1..m\}$ be the target ontology concept embeddings.
The optimal transport problem defined in Equation (\ref{eq:optimaltransport}) requires as input
ground costs $\mathbf{C}=[c(\emb{x}_{i}, \emb{y}_{j})]_{i, j}$ and probability distributions
$p(\emb{x}_{i})$ and $q(\emb{y}_{j})$.
For label embedding spaces, we use the Euclidean distances as the ground costs. 
For the probability distributions, we estimate non-uniform weights
using the shortest distances between source and target concept embeddings.
In particular, for a source point $\emb{x}_{i}\in \mathbf{X}$, let $d_{i} = \min_{j=1}^{m} c(\emb{x}_{i}, \emb{y}_{j})$
be the shortest distance from $\emb{x}_{i}$ to all embedding points in the target space. The distribution
$p(\emb{x}_{i})$ will be inversely proportional to $d_{i}$ for $i=1..n$. In other words, the greater the shortest
distance from a source point to all target points, the less the weight associated with the source concept.
Similarly, we estimate non-uniform probability distribution
$q(\emb{y}_{j})$ associated with the target concepts using the shortest distances from target points
to source points. 

The solution, 
$\mathbf{T}=[T(\emb{x}_{i}, \emb{y}_{j})]_{i, j}$, $i=1..n, j=1..m$, is 
a coupling matrix between every source and target embedding point. 
We test the following two methods for deriving candidate matchings from the coupling matrix:
\begin{itemize}
	\item \textbf{Mutual Nearest Neighbor (MNN):} for a $\emb{x}_{p}\in$$\{\emb{x}_{i}\in \R^{d}, i=1..n\}$, 
	find $\emb{y}_{q}\in$$\{\emb{y}_{j}\in \R^{d}, j=1..m\}$, such that, $T(\emb{x}_{p}, \emb{y}_{q})$
	$=$ $\max\{T(\emb{x}_{p}, \emb{y}_{j}), j=1..m\}$ and $T(\emb{x}_{p}, \emb{y}_{q})$
	$=$ $\max\{T(\emb{x}_{i}, \emb{y}_{q}), i=1..n\}$.
	\item \textbf{Top-K Targets (TopK):} for a $\emb{x}_{p}\in$$\{\emb{x}_{i}\in \R^{d}, i=1..n\}$, 
	find $k$ targets $\{\emb{y}_{q_{1}}, \emb{y}_{q_{2}}, ..,\emb{y}_{q_{k}}\}\subset$
	$\{\emb{y}_{j}\in \R^{d}, j=1..m\}$, such that, $T(\emb{x}_{p}, \emb{y}_{q_{z}})$
	$\geq$ $\max\{T(\emb{x}_{p}, \emb{y}_{j}), j\neq q_{1}..q_{k}\}$, for $z=1..k$.
\end{itemize}

\example{\emph{
 In Figure \ref{fig:OT-example}(a), the shortest distances from the source concepts to the targets concepts are
 $d_{\entity{Document}}=0.35$, $d_{\entity{Writer}}=0.37$, and $d_{\entity{Topic}}=0.58$. Taking inverses of the distances
 and normalizing them gives rise to a non-uniform source distribution $\mu=\{p(\entity{Document})=0.39, 
 p(\entity{Write})=0.37, p(\entity{Topic})=0.24\}.$ We can estimate the target probability distribution in the same way.
 After solving the optimal transport problem, we obtained
 the coupling matrix for the optimal transportation as shown in Figure \ref{fig:OT-example}(b). 
 By applying MNN, we obtain the following correspondences: 
 \entity{Document}\corresp \entity{Accepted\_Paper}, \entity{Write}\corresp \entity{Author},
 \entity{Topic}\corresp \entity{Paper}. By applying TopK (K=2), we obtain a 
 different set of correspondences each of which contains
 a set of potential target concepts for each source as follows:\\
  \entity{Document}\corresp \{\entity{Accepted\_Paper}, \entity{Subject\_Area}\}, \\
  \entity{Write}\corresp \{\entity{Author}, \entity{Paper}\},\\
  \entity{Topic}\corresp \{\entity{Paper}, \entity{Subject\_Area}\}.
}}

Our experimental results (see Section \ref{sec:experimental-results}) 
demonstrated 
that the global matching candidates (through MNN and TopK) outperformed most of 
the SOTA systems in terms of the \emph{recall} metrics. 
To improve its overall F1 measures, we develop refinement steps presented in next section.

%
%
\section{Refining Matching Candidates by Local Wasserstein Distances}
\label{sec:refining-matching}

We refine the candidate matchings by using Wasserstein distances between the 
local contexts of source and target ontology elements.
The local context of an element contains the triples
involving the element, including  the \entity{subClassOf} relationships connecting to the element's 
parents and children,
\entity{object} property, and \entity{datatype} property. 
For each matching candidate, we retrieve the local contexts of the source and target concepts. 
We then compute the local Wasserstein distance (localWD) between the local contexts by the following steps. First, compute
the pairwise Wasserstein distance (pairWD) between each pair of triples in the contexts.
The ground costs for computing the pairWDs are the Euclidean distances between 
the embeddings of the individual elements in the triples. Second, 
compute the local Wasserstein distance between the contexts by using
the pairWDs as ground costs. 

\begin{figure}[t]
	\centering
	\includegraphics[width=1\textwidth]{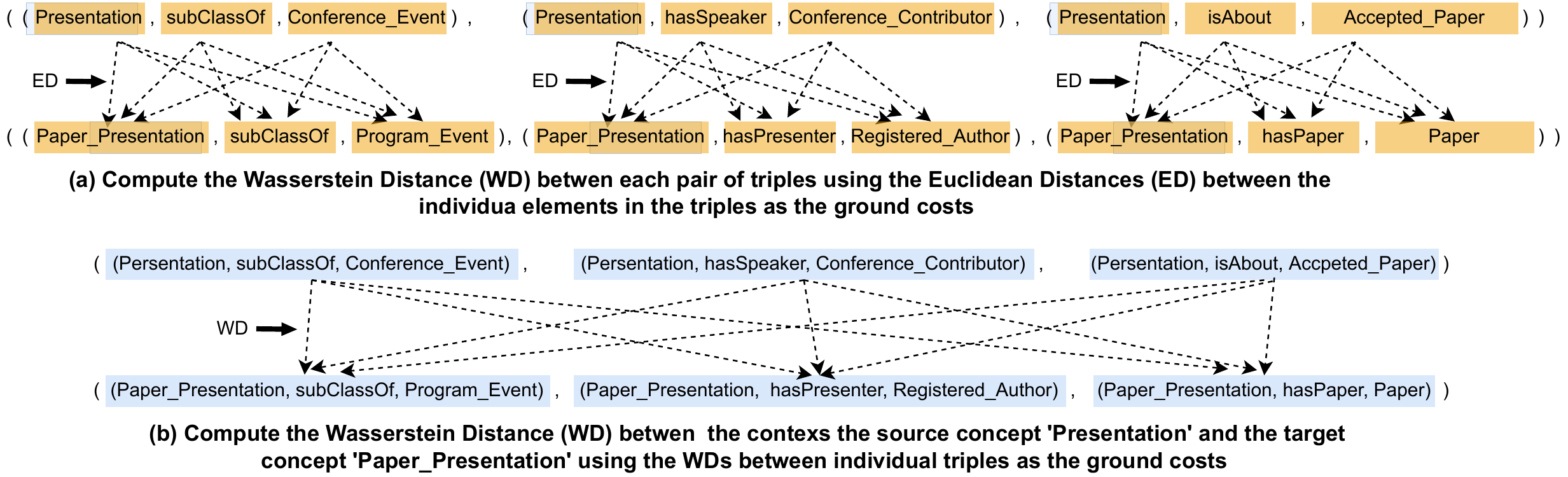}
	\caption{Illustration of computing local Wasserstein Distance (localWD) between the local
	contexts of a source concept `\entity{Presentation}' and a target concept `\entity{Paper\_Presentation}'.
The computation starts with computing the pairwise Wasserstein distance (pairWD) between each pair of triples in the contexts (a).
It then computes the localWD using the pairWDs as the ground costs (b).}
	\label{fig:wd_between_contexts}
\end{figure}

\example{
	\emph{
		Figure \ref{fig:wd_between_contexts}
		illustrates the two-step procedure for computing the localWD for a matching candidate 
		$\entity{Presentation}\corresp \entity{Paper\_Presentation}$. We first extract the local context of the 
		source concept \entity{Presentation} as a set of triples, $context(\entity{Presentation})=\{S_1, S_2, S_3\}$, as follows: \\
		$S_1$:(\entity{Presentation}, \entity{subClassOf}, 
		\entity{Conference\_Event}), \\
		$S_2$:(\entity{Presentation}, \entity{hasSpeaker}, 
		\entity{Conference\_Contributor}), \\
		$S_3$:(\entity{Presentation}, \entity{isAbout}, 
		\entity{Accepted\_Paper}).	\\
		Similarly, we extract the local context of the 
		target concept \entity{Paper\_Presentation} as a set of triples, 
		$context(\entity{Paper\_Presentation})=\{T_1, T_2, T_3\}$, as follows: \\
		$T_1$:(\entity{Paper\_Presentation}, \entity{subClassOf}, 
		\entity{Program\_Event}), \\
		$T_2$:(\entity{Paper\_Presentation}, \entity{hasPresenter}, 
		\entity{Registered\_Author}), \\
		$T_3$:(\entity{Paper\_Presentation}, \entity{hasPaper}, 
		\entity{Paper}). \\
		Given these two local contexts, we compute the pairwise WDs between the two sets of triples
		as illustrated in Figure \ref{fig:wd_between_contexts}(a) (where only
		$S_1\corresp T_1$, $S_2\corresp T_2$, and $S_3\corresp T_3$ are shown).
		Finally, we use the pairWDs as ground costs to compute the localWD between 
		$context(\entity{Presentation})$ and $context(\entity{Paper\_Presentation})$ as
		illustrated in Figure \ref{fig:wd_between_contexts}(b).
}}

After computing the local WDs for all matching candidates, we refine the candidate 
matchings using the localWDs as a main factor. We describe the set of  
experiments and results in next section. 

%
%
\section{Experiment and Results}
\label{sec:experimental-results}

\textbf{Setting Up.} We name the process of deriving matching candidates from global coupling matrix
as \emph{OTMapOnto\_}\texttt{global}. We add a suffix \texttt{\_mnn} or \texttt{\_topK} to it
to indicate whether the candidates are derived by mutual nearest neighbors or top-K targets. In our experiments,
we derive a large set of candidates by top-20 targets for refinement (thus, \emph{OTMapOnto-}\texttt{global\_top20} 
in the following tables illustrating the results). 
We name the process of 
refining matching candidates through local Wasserstein distances as 
\emph{OTMapOnto\_}\texttt{refinement}. We evaluate these processes on the Conference track
\footnote{http://oaei.ontologymatching.org/2021/conference/index.html} and 
MSE benchmark\footnote{https://github.com/EngyNasr/MSE-Benchmark}
in the OAEI Campaign. The code and Jupyter notebooks 
for the experiments is here\footnote{https://github.com/anyuanay/otmaponto\_django}.

In the refinement process, we create interactions among 
localWDs, string-based distances, and embedding-based Euclidean distances. 
The interactions are performed by multiplication. We then examine 
any enhancements brought by the localWDs in comparison to only string-based distances, embedding-based
Euclidean distances, and their interactions. We compare the following cases:
\begin{itemize}
	\item String-based Levenshtein distances/similarities (string-based distance)
	\item Interactions between the string-based distances and localWDs (string-context-distance)
	\item Euclidean distances between the averaged embeddings of labels 
	\item Interactions between the Euclidean distances and localWDs
	\item Wasserstein distances between the embeddings of labels
	\item Interactions between the label Wasserstein distances and localWDs
	\item Interactions among string-based distances, Euclidean distances, label Wasserstein distances, and localWDs 
\end{itemize} 

We converted each distance metric $x$, 
to a similarity metric in $[0,1]$ by taking $e^{-x}$. We run through thresholds from 0 to 1 in a step of 0.01
to find the best performance. Our experimental results show that the interactions between the string-based distances and 
localWDs (\emph{string-context-distance})
achieve the best performance among all distance metrics. In the following tables, 
we specifically report on the values related to the \emph{string-context-distance} metric.  

\noindent
\textbf{Evaluation on Conference Track.} There are 21 matching cases among 7 conference ontologies. 
We adopt the main reference alignment \textbf{rar2} with class only case (\textbf{M1}) for evaluation. 
Table \ref{tab:rar2-M1} contains
the \textbf{rar2-M1} results of the OAEI 2021 Campaign, plus the \textbf{rar2-M1} results of 
\emph{OTMapOnto\_}\texttt{global\_mnn}, \emph{OTMapOnto\_}\texttt{global\_top20}, and 
\emph{OTMapOnto\_}\texttt{refinement}. The results show that the matching candidates derived from the global 
coupling matrix through either MNN or Top-K achieved high recall but low precision. 
With the refinement, \otmap 
achieved the best precision and a compatible F1 measure compared to the best tools in the campaign. 
\begin{table}[!th]
	\begin{center}
		\begin{tabular}{|l|r|r|r|r|}
			\hline
			\textbf{Matcher} & \textbf{Threshold} & \textbf{Precision} & \textbf{Recall} & \textbf{F1} \\
			\hline
			AML & 0 & 0.76 & 0.66 & \cellcolor[HTML]{c0c0c0}\textbf{0.71} \\
			\hline
			\emph{OTMapOnto-}\texttt{refinement} & 0.30 & \cellcolor[HTML]{C0C0C0}\textbf{0.89} & 0.59 & 0.70 \\
			\hline
			GMap & 0 & 0.70 & 0.67 & 0.68 \\
			\hline
			LogMap & 0 & 0.78 & 0.60 & 0.68 \\
			\hline
			Wiktionary & 0 & 0.78 & 0.57 & 0.66 \\
			\hline
			FineTOM & 0 & 0.73 & 0.58 & 0.65 \\
			\hline
			TOM & 0.86 & 0.82 & 0.52 & 0.64 \\
			\hline
			ATMatcher & 0 & 0.69 & 0.59 & 0.64 \\
			\hline
			ALOD2Vec & 0.29 & 0.79 & 0.54 & 0.64 \\
			\hline
			edna & 0 & 0.82 & 0.51 & 0.63 \\
			\hline
			LogMapLt & 0 & 0.78 & 0.52 & 0.62 \\
			\hline
			StringEquiv & 0 & 0.83 & 0.48 & 0.61 \\
			\hline
			LSMatch & 0 & 0.83 & 0.48 & 0.61 \\
			\hline
			AMD & 0 & 0.81 & 0.48 & 0.60 \\
			\hline
			KGMatcher & 0 & 0.83 & 0.45 & 0.58 \\
			\hline
			Lily & 0.24 & 0.62 & 0.52 & 0.57 \\
			\hline
			\emph{OTMapOnto-}\texttt{global\_mnn} & 0 & 0.24 & 0.73 & 0.36 \\
			\hline
			\emph{OTMapOnto-}\texttt{global\_top20} & 0 & 0.007 & \cellcolor[HTML]{c0c0c0}\textbf{0.96} & 0.013 \\
			\hline
		\end{tabular}
	\end{center}
	\caption{\label{tab:rar2-M1} Evaluation on OAEI 2021 conference track ordered by f1 in descending order. 
	The shaded value is the best in its category.}
\end{table}

\noindent
\textbf{Evaluation on MSE Benchmark.} The MSE benchmark contains 3 matching cases between 3 materials
science and engineering ontologies: MaterialInformation, MatOnto, and EMMO (European Material
Modeling ontology). We downloaded the 
two leading ontology matching systems AML\footnote{https://github.com/AgreementMakerLight/AML-Project}
and LogMap\footnote{https://github.com/ernestojimenezruiz/logmap-matcher} for comparison. The results are presented in 
Table \ref{tab:mse}. The table 
shows \emph{OTMapOnto\_}\texttt{refinement} achieved the best F1 performance 
for all 3 test cases.
\begin{table}[!th]
	\begin{center}
		\begin{tabular}{|c|l|r|r|r|}
			\hline
			\textbf{Test Case} & \textbf{Matcher} & \textbf{Precision} & \textbf{Recall} & \textbf{F1} \\
			\hline
			\multirow{3}{*} 
			{Case 1} & \emph{OTMapOnto-}\texttt{refinement} & 0.78 & 0.30 & \cellcolor[HTML]{C0C0C0}\textbf{0.44} \\
			\cline{2-5}
			& \emph{OTMapOnto-}\texttt{global\_mnn} & 0.23 & 0.39 & 0.29 \\
			\cline{2-5}		 
			& AML & 0.80 & 0.17 & 0.29 \\
			\cline{2-5}
			 & LogMap & \cellcolor[HTML]{C0C0C0}\textbf{1.00} & 0.04 & 0.08 \\
			 \cline{2-5}
			 & \emph{OTMapOnto-}\texttt{global\_top20} & 0.001 & \cellcolor[HTML]{C0C0C0}\textbf{0.78} & 0.002 \\
			\hline
			\hline
			\hline
			\multirow{3}{*} 
			{Case 2} & \emph{OTMapOnto-}\texttt{refinement} & 0.31 & 0.54 & \cellcolor[HTML]{C0C0C0}\textbf{0.39} \\
			\cline{2-5}
			& \emph{OTMapOnto-}\texttt{global\_mnn} & 0.54 & 0.26 & 0.35 \\
			\cline{2-5}		 
			& AML & 0.82 & 0.21 & 0.34 \\
			\cline{2-5}
			& LogMap & \cellcolor[HTML]{C0C0C0}\textbf{0.87} & 0.20 & 0.32 \\
			\cline{2-5}
			& \emph{OTMapOnto-}\texttt{global\_top20} & 0.01 & \cellcolor[HTML]{C0C0C0}\textbf{0.57} & 0.02 \\
			\hline
			\hline
			\hline
			\multirow{3}{*} 
			{Case 3} & \emph{OTMapOnto-}\texttt{refinement} & \cellcolor[HTML]{C0C0C0}\textbf{0.98} & 0.87 & \cellcolor[HTML]{C0C0C0}\textbf{0.92} \\
			\cline{2-5} 
			& AML & 0.96 & 0.87 & 0.91 \\
			\cline{2-5}
			& LogMap & 0.93 & 0.84 & 0.88 \\
			\cline{2-5}
			& \emph{OTMapOnto-}\texttt{global\_mnn} & 0.67 & 0.86 & 0.75 \\
			\cline{2-5}
			& \emph{OTMapOnto-}\texttt{global\_top20} & 0.007 & \cellcolor[HTML]{C0C0C0}\textbf{0.97} & 0.014 \\
			\hline
		\end{tabular}
	\end{center}
	\caption{\label{tab:mse} Evaluation on MSE benchmark ordered by f1 in descending order in each case.
	The shaded value is the best in its category.}
\end{table} 

%
%
%
\section{Discussion}
\label{sec:discussion}

We name our system as \otmap \cite{OTMapOnto}. Our experimental results showed 
\otmap system achieved promising results. 
Most significantly, for the MSE benchmarks, \otmap with refinements
outperformed the two leading matching systems that have been consistently the top 2 in previous OAEI campaigns
in many tracks. \otmap also outperformed almost all of the systems participating in the OAEI 2021 Conference track.
This exploratory study provided positive answers to our exploration questions listed in Introduction.
First, Wasserstein distance is effective in capturing semantic similarity between ontologies. 
Second, the coupling matrix returned by the optimal transport solver contains 
most of the correct matchings but with many spurious ones. Both sets of candidates derived 
through mutual nearest neighbors (MNN) and Top-K targets (TopK)
achieved the best recall results. We also observed this phenomenon in several tasks in the OAEI 2021 Campaign 
\cite{OTMapOnto}, where the results had lower precision and 
higher recall compared to other systems.   
Finally, we found that using local Wasserstain distances between the contexts of the source and target
concepts of a candidate greatly helps filter out many false positives. 
The final overall performance metrics are better or compatible to the SOTA results. 

In discussing these results, it is important to note that this exploratory study
only considered the embeddings of concept labels generated by a
pre-trained model. In future work, we will develop ontology embeddings
capturing additional ontology information including structural and logical components. 
For large scale ontologies, the $n\times m$ matrices associated with optimal transport
are unscalable. It is quite evident that we need to break down the ontologies into smaller chunks for computing
the optimal transport couplings. In moving forward, we will first partition the embeddings into clusters. 
Using Wasserstein distances to measure the similarities of pairs of clusters,  we will aim to find
candidate matchings from the pairs of clusters that are most similar.

%
%
%
\section{Related Work}
\label{sec:related-work}

AML \cite{augmenting-ontology-alignment} and LogMap \cite{LogMap}
are two leading classical matching systems based on symbolic structures. 
Nkisi-Orji et al. in \cite{ontology-alignment-embedding-random-forest} 
developed an embedding-based supervised machine learning method.
Kolyvakis et al. in \cite{biomedical-ontology-alignment} presented an approach that 
applied the Stable Marriage algorithm for deriving candidate matchings
based on the cosine similarity between retrofitted embedding vectors. 
Chen et al. in \cite{augmenting-ontology-alignment}
proposed a distantly supervised method that combines embedding-based extensions
with classical systems such as AML and LogMap. More recently, 
He et al. in \cite{BERTMap} described a transformer-based ontology matching system, BERTMap.
Alvarez-Melis et al. in \cite{pmlr-v108-alvarez-melis20a}
proposed a method that first encodes 
hierarchical data in hyperbolic spaces and then applies
optimal transport to the hyperbolic embeddings for deriving correspondences.
However, the performance for ontology matching is limited when only the hyperbolic embeddings of 
ontological hierarchies were considered.

%
%
\section{Conclusion}
\label{sec:conclusion}

We explored the effectiveness of the Wasserstein distance metric defined by optimal transport
process for ontology matching. 
Our study showed Wasserstein distance is effective in measuring the similarity between 
ontologies. The experimental results showed the
Wasserstein distance-based approach outperformed almost all the cases in the test data.
We plan to test the approach on a wide range of ontology matching applications. 
In addition, for source and target ontology embedding spaces
without `registration', that is, they do not have well-defined 
ground distance between them,  we will extend to Gromov Wasserstein distance metric \cite{Gromov-W}
which measures how distances between pairs of concepts are matched across ontologies.

%
%

\bibliographystyle{splncs04}

\end{document}